\documentclass[lettersize,journal]{IEEEtran}
\usepackage{amsmath,amsfonts}
\usepackage{algorithm}
\usepackage{array}
\usepackage[caption=false,font=normalsize,labelfont=sf,textfont=sf]{subfig}
\usepackage{textcomp}
\usepackage{stfloats}
\usepackage{url}
\usepackage{verbatim}
\usepackage{graphicx}

\usepackage{graphicx}
\usepackage{amsmath}
\usepackage{amssymb}
\usepackage{booktabs}
\usepackage{amsfonts}
\usepackage{textcomp}
\usepackage{xcolor}
\usepackage{breqn}
\usepackage{multirow}
\usepackage{array}
\usepackage{threeparttable}
\usepackage{algorithm}
\usepackage{algpseudocode}
\usepackage{algorithmicx}
\usepackage{tabularx}

\usepackage{balance}

\begin{document}
\title{Few-Shot Continual Learning via Flat-to-Wide Approaches}
\author{Muhammad Anwar Ma'sum, Mahardhika Pratama,~\IEEEmembership{Senior member, ~IEEE}, Edwin Lughofer, Lin Liu, Habibullah, Ryszard Kowalczyk}
\markboth{Journal of \LaTeX\ Class Files,~Vol.~14, No.~8, August~2021}%
{Shell \MakeLowercase{\textit{et al.}}: A Sample Article Using IEEEtran.cls for IEEE Journals}


\maketitle

\begin{abstract}
Existing approaches on continual learning call for a lot of samples in their training processes. Such approaches are impractical for many real-world problems having limited samples because of the overfitting problem. This paper proposes a few-shot continual learning approach, termed FLat-tO-WidE AppRoach (FLOWER), where a flat-to-wide learning process finding the flat-wide minima is proposed to address the catastrophic forgetting problem. The issue of data scarcity is overcome with a data augmentation approach making use of a ball generator concept to restrict the sampling space into the smallest enclosing ball. Our numerical studies demonstrate the advantage of FLOWER achieving significantly improved performances over prior arts notably in the small base tasks. For further study, source codes of FLOWER, competitor algorithms and experimental logs are shared publicly in \url{https://github.com/anwarmaxsum/FLOWER}.
\end{abstract}

\begin{IEEEkeywords}
Continual Learning, Few-Shot Learning, Incremental Learning, Lifelong Learning
\end{IEEEkeywords}

\section{Introduction}\label{sec1}
Existing studies on continual learning (CL) mostly focus on a solution of the catastrophic forgetting (CF) problem where old parameters of a deep neural network are overwritten when learning new tasks without relying on any information of old samples thus losing its generalization power of old tasks. Three approaches, namely regularization-based approach, structure-based approach, memory-based approach, have been well-established in the literature \cite{Parisi2019ContinualLL} where some of the above mentioned approaches have successfully tackled the class-incremental learning problem \cite{Ven2019ThreeSF}, i.e., a single head scenario without any task IDs, more challenging than the task-incremental learning problem. However, these approaches suffer from the over-fitting problem in the case of limited samples. Hence, they are unfit for many real world problems such as a slow-speed manufacturing process where the problem of data scarcity is present. This drawback also does not coincide with the natural learning where human being can learn from very few samples \cite{Finn2017ModelAgnosticMF}. Our interest here is to attack the problem of few-shot continual learning \cite{Tao2020FewShotCL} where each task carries very few samples in the N-way K-shot setting, i.e., each task comprises N classes where each task presents K samples per class. Such a case requires a model to handle not only a sequence of tasks without the CF problem but also to avoid the over-fitting problem due to scarce samples.

The few-shot continual learning approach is pioneered by the work in \cite{Tao2020FewShotCL} where the concept of neural gas (NG) is adopted. Old nodes of the NG structure are stabilized to avoid the CF problem while expanding the NG structure to adapt to new concepts. \cite{Mazumder2021FewShotLL} presents a few-shot continual learning algorithm using the concept of session trainable parameters, deemed unimportant for previous tasks to alleviate the CF problem. The training process is governed by a triplet loss function and classification is performed using the nearest prototype-based classification approach. The vector quantization concept in the deep embedding space is successfully implemented in \cite{Chen2021IncrementalFL} to deal with both regression and classification problems in few-shot continual learning environments. A continually evolved classifier (CEC) is proposed in \cite{Zhang2021FewShotIL} using a graph attention network (GAT) to adapt the classifier's weights. \cite{Shi2021OvercomingCF} introduces the concept of flat learning, finding flat regions in the base learning task to be maintained in the few-shot learning task. However, it relies on a memory storing exemplar sets of previous tasks. The use of memory might lead to most samples of few-shot learning tasks to be kept in the memory. It heavily focuses on stability of base tasks but ignores plasticity of new tasks.

This paper presents a FLat-tO-WidE appRoach (FLOWER) for few-shot continual learning problems. FLOWER is configured as a prototypical network \cite{Snell2017PrototypicalNF} driven by the prototypical loss. FLOWER puts forward the flat-to-wide learning idea \cite{Shi2021OvercomingCF,Cha2021CPRCR}, to discover flat-wide minimum regions. It minimizes the losses of noise-perturbed parameters of the feature extractor over $M$ runs. The flat-wide parameters are maintained in the subsequent few-shot learning tasks by constraining feature extractor parameters to prevent the catastrophic forgetting problem. Our approach differs from \cite{Shi2021OvercomingCF} heavily focusing on the flat minimum regions resulting in over-dependence on the base learning task. In addition, no memory is imposed by FLOWER, and the wide learning concept is integrated as an attempt to drive flat regions to be wide, thus scaling well for large-scale problems. The concept of ball generator is integrated in FLOWER to prevent the over-training problem due to scarce samples in few-shot learning tasks.

A feature-space data augmentation approach is applied in a sequence of few-shot learning tasks. The data augmentation strategy follows the ball generator concept \cite{Sun2021MEDAMW} where the sampling space is limited to the smallest enclosing ball of few-shot samples. A generator loss is incorporated such that generated samples are adjacent to the center of its own class and away to the centre of other classes. The projection-based memory aware synapses method is implemented to avoid classifier's parameters being overwritten when embracing new few-shot tasks. That is, the memory aware synapses regularization approach is combined with the classifier's projection concept of \cite{Cha2021CPRCR} and the clamping procedure inducing flat-wide local optimum regions scaling well for large-scale continual learning tasks. The main network and the ball generator are trained to minimize a joint loss function of the prototypical loss and the generator loss added with the regularizer. A constraint is implemented to feature extractor parameters where they are clamped if going beyond the flat-wide regions \cite{Shi2021OvercomingCF}. This approach also prevents the over-fitting problem because only few classifier parameters are adapted while leaving feature extractor parameters in a fixed range.

Five major contributions are proposed: 1) this paper proposes flat-to-wide approach (FLOWER) for few-shot continual learning synergizing the flat learning concept and the wide learning concept. The flat-wide minima are enforced to address the CF problem. This enables improved plasticity when learning new tasks as well as maintains decent performances with small base tasks. Note that the good average performance of \cite{Shi2021OvercomingCF} often comes at the cost of poor performances on new tasks and small base tasks; 2) the concept of ball generator is proposed to perform the feature space data augmentation as a solution of data scarcity. This approach is modified from \cite{Sun2021MEDAMW} designed for texts and non-continual learning problems; 3) new loss functions unifying the flat learning loss, wide learning loss and ball-generator loss are derived for both base learning task and few-shot learning tasks leading to an end-to-end training process; 4) rigorous theoretical study is performed to confirm the convergence of FLOWER where its losses converge toward zero; 5) the efficacy of FLOWER has been validated and compared with state-of-the-art algorithms where FLOWER delivers improved performances and works well with the small base tasks. \textit{Source codes of FLOWER are placed in \url{https://github.com/anwarmaxsum/FLOWER} to assure convenient reproductions of our numerical results and further studies.}.

\section{Related Works}
\subsection{Continual Learning}
There exist three variants of CL approaches notably to attack the CF problem \cite{Parisi2019ContinualLL}: regularization-based approach, memory-based approach and architecture-based approach. The regularization-based approach relies on a penalty term for important parameters of previous tasks making sure no significant deviations when handling new tasks \cite{kirkpatrick2016overcoming,Zenke2017ContinualLT,Aljundi2018MemoryAS,Paik2020OvercomingCF,Mao2021ContinualLV,Cha2021CPRCR,Li2018LearningWF}. The performance of regularization-based approaches is usually inferior to other two approaches and often call for the presence of task IDs, thus performing poorly in the Class-Incremental Learning (CIL).  The structure-based approach applies a network expansion strategy combined with a parameter isolation strategy to overcome the CF problem \cite{Rusu2016ProgressiveNN,Yoon2018LifelongLW,Zoph2017NeuralAS,Li2019LearnTG,Xu2021AdaptivePC,Ashfahani2022UnsupervisedCL,Pratama2021UnsupervisedCL,Rakaraddi2022ReinforcedCL}. In addition to expensive computational costs, above-mentioned works depend on task IDs and task boundaries, thus being inapplicable to deal with few-shot CIL problems.  Although the data-driven approach is fast to adapt to changes and does not require any task IDs and task boundaries, it does not guarantee an optimal structure. The memory-based approach stores a subset of old samples in an episodic memory for experience replay when dealing with new tasks and often offers strong performances in the CIL problems \cite{Rebuffi2017iCaRLIC,LopezPaz2017GradientEM,Chaudhry2019EfficientLL,Chaudhry2021UsingHT,Buzzega2020DarkEF,Dam2022ScalableAO,VinciusdeCarvalho2022ClassIncrementalLV}. Although these approaches perform strongly for the CIL problems, the memory-based approach is impractical for few-shot CL problems because of the limited sample constraint, i.e., it is akin to the retraining case undermining the CL spirit. 

\subsection{Few-Shot Continual Learning}
Few-shot CL problem aims to resolve the data-scarcity issue in addition to the CF problem. \cite{Tao2020FewShotCL} is seen as a pioneering approach in this area where the concept of neural gas is adopted to address new concepts without compromising already seen experiences. \cite{Mazumder2021FewShotLL} puts forward session trainable parameters to avoid the over-training problem while using the triplet-loss based metric learning. \cite{Chen2021IncrementalFL} applies the vector quantization concept and has been validated for regression as well as classification problems. \cite{Zhang2021FewShotIL} makes use of graph attention network (GAT) to control the adaptive mechanisms of new tasks. The concept of flat learning is proposed in \cite{Shi2021OvercomingCF} for the few-shot CL problems. This approach is, however, dependent on a memory which should not exist for the few-shot CL problems, i.e., all samples can be stored in the memory thus becoming equivalent to retraining approaches. FLOWER also differs from \cite{Shi2021OvercomingCF} where the ball generator concept is integrated for the feature space data augmentation mechanism to better address the data scarcity problem while introducing the wide learning concept to scale up to large-scale tasks. 

\section{Problem Formulation}
A few-shot continual learning problem is defined as a learning problem of a sequence of fully-labelled tasks $\mathcal{T}_1,\mathcal{T}_k,...,\mathcal{T}_K$ where $k\in\{1,...,K\}$ denotes the number of tasks unknown before a process runs. Each task carries $N_k$ pairs of training samples $\mathcal{T}_{k}=\{x_i^{k},y_i^{k}\}_{i=1}^{N_k}$ where $x_i\in\mathcal{X}$ stands for an input image and $y_i\in\mathcal{Y}$ denotes its corresponding class label. All data samples are drawn from the same domain $\mathcal{X}\times\mathcal{Y}\in\mathcal{D}$. Each task features the same image size, but possesses disjoint target classes. Suppose that $Y_k$ labels a class set of a $k-th$ task and $Y_{k-1}$ denotes a class set of a $k-1$ task, $\forall k,k-1, Y_k\cap Y_{k-1}=\emptyset$. The few-shot continual learning problem is commonly formulated as the class-incremental learning problem rather than the task-incremental learning problem \cite{Ven2019ThreeSF} where a class prediction is inferred by a single-head network structure and a pair of data samples of every task arrive without the task IDs $t_i^{k}$. 

$\mathcal{T}_1$ is a base learning task possessing a reasonably large number of samples while the remainder of the tasks is a few-shot learning task lacking of samples, i.e., each few shot learning task is set in the $N$ way $K$ shot configuration where it has $N$ target classes with $K$ samples per classes. In addition, a task $\mathcal{T}_{k>1}$ is only accessible at the $k-th$ session and completely discarded once seen. This constraint causes the catastrophic interference problem where previously valid parameters are over-written when learning a new task. The memory-based approach is  unfeasible for the few-shot continual learning problems because it may lead to the storage of all past samples due to the data scarcity constraint, a retraining case. Hence, a few-shot continual learner is supposed to not only accumulate knowledge from already seen experiences but also present efficient sample utilization to avoid the over-fitting problem.  

Our few-shot continual learner is defined as $g_{W_{C}}(f_{W_{F}}(.))$ where $f_{W_{F}}(.)$ stands for a feature extractor with parameters $W_{F}$ while $g_{W_{C}}(.)$ denotes a fully-connected network with parameters $W_{C}$. Specifically, it is configured as a prototypical network \cite{Snell2017PrototypicalNF} where the final prediction is expressed as follows:
\begin{equation}
    P(\hat{y}=c|x;W_{C},W_{F})=\frac{\exp{(-d(g_{W_C}(f_{W_F}(x)),p_c))}}{\sum_{c\in M}\exp{(-d(g_{W_C}(f_{W_F}(x)),p_c))}}   
\end{equation}
$c^*=\arg\max_{c\in M}P(\hat{y}=c|x;W_{C},W_F)$ stands for a predicted class label and $M$ denotes the number of classes seen thus far. $d(.)$ is a distance metric implemented as the $L_2$ distance metric in this paper. $p_{c}=\frac{1}{|Y_1|}\sum_{(y)\in Y_1}f_{W_{F}}(x_i)$ is the prototype of the $c-th$ class. The learner is tasked to perform well across all tasks and minimizes $\frac{1}{K}\sum_{k=1}^{K}\mathcal{L}_{k}$ and $\mathcal{L}_{k}\triangleq\mathbb{E}_{(x,y)\backsim\mathcal{T}_{k}}[\mathcal{L}(g_{W_{C}}(f_{W_{F}}(x)),y)]$ where $\mathcal{L}(.)$ is a loss function of interest where a cross-entropy loss of the prototypical network is adopted here.

\begin{equation}\label{LCE}
    \mathcal{L}_{ce}=-\sum_{i}\sum_{c}\mathbf{1}(y=c)\frac{\exp{(-d(g_{W_C}(f_{W_F}(x_i)),p_c))}}{\sum_{c\in M}\exp{(-d(g_{W_C}(f_{W_F}(x_i)),p_c))}}
\end{equation}
where $\mathbf{1}(y=c)$ is an indicator function returning 1 if $y=c$. The inference adopts the nearest-mean classification strategy making use of a class label of the nearest prototype.
\section{The Flat-to-Wide Approach (FLOWER)}
The few-shot continual learning problem consists of two learning stages: a base learning phase and a sequence of few-shot learning phases.
\subsection{Base Learning Phase ($k=1$)}
\begin{figure}[!t]
 \centering
 \includegraphics[scale=0.7]{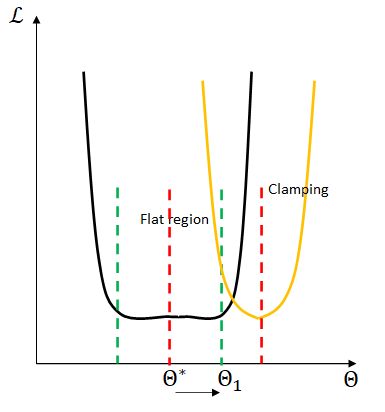}
 \caption{Flat Continual Learning: during the base learning phase $k=1$, FLOWER finds the network parameters $\Theta^*$ located in the flat local optimum region. These parameters might shift to non-flat local minima regions when learning the few-shot learning tasks $k>1$ via $\Theta^*\rightarrow\Theta_1$. Clamping is performed for $\Theta_1$ to stay around a flat local optimum region.}
 \label{FlatCL}
 \end{figure}

We apply the concept of b-flat-wide local minima where, for the parameters of interests in the flat-wide region, i.e., the feature extractor parameters $W_{F}$ the losses are still minimized leading to well separations of the classes of interest and robustness against the catastrophic forgetting problem \cite{Shi2021OvercomingCF}. Once found in the base learning phase, these parameters are only tuned within this region during the few-shot learning phases. In other words, it is sufficient to adjust the network parameters within the flat-wide interval to avoid the over-training problem. A definition of the b-flat-wide local minima is formalised as:

\noindent\textbf{Definition 1} \cite{Shi2021OvercomingCF}: given a real-valued loss function $\mathcal{L}(.)$, for any $b>0$ and network parameters $\Theta=\{W_{C},W_{F}\}$, the b-flat-wide local minima $\Theta^*$ follow the following condition:
\begin{itemize}
    \item Condition 1: for $\mathcal{L}(x;\Theta^*+\epsilon)$, where $-b\leq\epsilon\leq+b$.
    \item Condition 2: there exist $c_1<\Theta^*-b$ and $c_2>\Theta^*+b$ s.t. $\mathcal{L}(x;\Theta)>\mathcal{L}(x;\Theta^*)$ where $c_1<\Theta<\Theta^*-b$ and $\Theta^*+b<\Theta<c_2$
\end{itemize}
These conditions are extremely hard to hold in reality and an approximation is carried out using a noise perturbation. That is, we run $M$ trials with different noise perturbations $\epsilon_j, -b\leq\epsilon_j\leq +b\quad\&\quad j\in M$, against network parameters $\Theta^*$. Under different noise perturbations, the network parameters within the estimated b flat-wide regions are supposed to return similar losses and to have small function values \cite{Shi2021OvercomingCF}. Fig. \ref{FlatCL} visualizes the flat learning paradigm. In a nutshell, the base learning phase is designed to minimize the following loss function.
\begin{equation}\label{LFLAT}
    \begin{split}
        \mathcal{L}_{k=1}=\frac{1}{M}\sum_{j=1}^{M}\mathcal{L}_{base}^{j}(g_{W_C}(f_{W_F+\epsilon_j}(x)),y)\\
        \mathcal{L}_{base}^{j}=\mathbb{E}_{(x,y)\backsim\mathcal{T}_{1}}[\mathcal{L}_{ce}(g_{W_C}(f_{W_F+\epsilon_j}(x)),y)+\\\mathcal{L}_{KL}(g_{W_{C}}(f_{W_F+\epsilon_j}(x)),P_{U})]+\lambda_1\mathbb{E}_{y\backsim Y_1}||p_c-p_c^*||_2
    \end{split}
\end{equation}
where $p_c,p_c^*$ stand for the prototype of the $c-th$ class before and after noise perturbations respectively. The first term of $\mathcal{L}_{base}$ is designed to unveil the b-flat-wide region while the second term prevents the prototype drift problem. $P_U$ is a uniform distribution and $\mathcal{L}_{KL}(.)$ promotes wide local minima via output's projections to uniform distributions. Note that \eqref{LFLAT} extends \cite{Shi2021OvercomingCF} where only the flatness is solicited and ignores the wide learning issue.

 The goal of \eqref{LFLAT} is to unveil a flat-wide region working well on all tasks, thus overcoming the catastrophic forgetting problem. It is done by sampling $\epsilon_j$ $M$ times to vary network parameter leading to $\mathcal{L}_{k=1}$ considered as a non-convex loss function \cite{Shi2021OvercomingCF}. It is an approximation of the expected loss $R(\Theta)=\mathbb{E}_{z,\epsilon}[\mathcal{L}_{t=1}]$, which is impossible to minimize in practise. $z_{k}$ is the $k-th$ data batch.
 
 \noindent\textbf{Assumption 1} (L-smooth risk function \cite{Shi2021OvercomingCF}): $R:\Re^{D}\rightarrow\Re$ is continuously differentiable and L-smooth $L>0$:
 \begin{equation}
     ||\nabla R(\Theta) - \nabla R(\Theta^{'})||_2\leq L||\Theta-\Theta^{'}||
 \end{equation}
 This assumption constraints the pace of the gradients w.r.t the parameter vectors. 
 
 \noindent\textbf{Assumption 2}: the expected loss function $R(\Theta)$ holds the following properties \cite{Shi2021OvercomingCF}: 
 \begin{itemize}
     \item $R$ is bounded below a scalar $R^*$ given $\{\Theta_k\}$ where $k$ indicates the $k-th$ data batch. It implies $R$ to be bounded by a minimum value $R^*$;
     \item For all $k\in\mathbb{N}$ and $j\in[1,M]$:
     \begin{equation}
         \mathbb{E}_{z_{k},\epsilon_{j}}\nabla\mathcal{L}_{base}^{j,k}=\nabla R(\Theta_k)
     \end{equation}
     where $\nabla\mathcal{L}_{base}^{j,k}$ is an unbiased estimation of $\nabla R(\Theta_{k})$ to simplify the proof; 
     \item given $m_1\geq0, m_2\geq0$, for all $k\in\mathbb{N}$ and $j\in[1,M]$:
     \begin{equation}
         \mathbb{V}_{z_k,\epsilon_j}[\nabla\mathcal{L}_{base}^{j,k}]\leq(m_1+m_2)||\nabla R(\Theta_k)||_{2}^{2}
     \end{equation}
     meaning that the variance of the gradient cannot be arbitrarily large.  
 \end{itemize}
 $\mathbb{E}_{z_{k},\epsilon_{j}}, \mathbb{V}[.]$ respectively denote the expectation w.r.t the joint distribution of random variables $z_k,\epsilon_j$ and the variance. These three conditions are reasonable in practise and applicable for convergence analysis. 
 
 \noindent\textbf{Assumption 3}: the learning rates $\lambda_k$ satisfy \cite{Shi2021OvercomingCF}
 \begin{equation}
     \sum_{k=1}^{\infty}\lambda_k=\infty, \sum_{k=1}^{\infty}\lambda_{k}^{2}<\infty
 \end{equation}
 which can be easily met due to $\lambda_k<1$ and a decreasing function w.r.t $k$. These assumptions establish a theorem.  
 
 \noindent\textbf{Theorem 1}: Given the assumptions 1-3 and $R$ is twice differentiable, we have:
 \begin{equation}
     \lim_{k\rightarrow\infty}\mathbb{E}[||\nabla R(\Theta_k)||_{2}^{2}]= 0 
 \end{equation}
 the convergence of the flat-wide minima strategy. \textit{A set of proofs are provided in the appendices}.

\subsection{Few-Shot Learning Phase ($k>1$)}
The few-shot learning phase of FLOWER is driven by a feature-space data augmentation approach to address the data scarcity issue while applying a combination of the flat and wide learning principles to address the CF problem without any memory. Network parameters, namely those of feature extractors, are adjusted in such a way to maintain the b-flat-wide local minima conditions to mitigate the over-training problems and the projection concept is implemented in the memory aware synapses (MAS) method \cite{Zenke2017ContinualLT} to induce wide local optimum. 
\subsubsection{Feature Space Data Augmentation} The concept of ball generator \cite{Sun2021MEDAMW} is adopted and operates in the feature space $z=f_{W_F}(x)\in\Re^{D}$ rather than the high-dimensional data space $x\in\Re^{w\times H\times c}$. The ball generator produces synthetic samples of the smallest enclosing ball of the $c-th$ class $\omega(Z_c)=\{C_c,\sigma_c\}$ where $C_c,\sigma_c$ denote the centroid and radius of the $c-th$ ball/class. Unlike \cite{Sun2021MEDAMW}, the ball generator is directly executed in the few-shot learning tasks where there exist the data scarcity problem rather than to that of the support set. A synthetic sample $\hat{z}\in\Re^{D}$ is generated:
\begin{equation}\label{synthetic}
    \hat{z}=C_c+u^{\frac{1}{D}}\sigma_c\frac{\phi}{||\phi||_2}
\end{equation}
where $u$ denotes a shifting constant sampled from a uniform distribution $[0,1]$ and $\phi\in\Re^{D}$ labels a random vector generated from a normal distribution $\phi\backsim\mathcal{N}(0,1)$. 

We feed a synthetic sample $\hat{z}$ to a transformation module $\hat{\hat{z}}=\kappa_{W_T}(\hat{z})\in\Re^{D}$ arranged as three fully connected layer network as per \cite{Sun2021MEDAMW}. The goal of the transformation module is to avoid the bias of synthetic samples. That is, a ball generator loss is introduced as follows:
\begin{equation}\label{LBALL}
    \mathcal{L}_{ball}=\sum_{\hat{\hat{z}}\in c_i,\hat{\hat{z}}\notin c_j}\lambda_{2}\max{\{0,d(\hat{\hat{z}},C_i)+r-d(\hat{\hat{z}},C_j)\}}
\end{equation}
where $r$ is a predefined margin and $\hat{\hat{z}}=\kappa_{W_T}(\hat{z})$ stands for a synthetically generated sample of the transformation module. $C_i, C_j$ respectively denote the centroid of the $i-th$ and $j-th$ ball or class. \eqref{LBALL} is inspired by the triplet loss of the metric learning aiming to pull the synthetic sample $\hat{\hat{z}}$ close to its centroid and to push it away from other centroids. $\lambda_2$ is a predefined constant steering the influence of the ball generator loss function. 
\subsubsection{Projection-based Memory Aware Synapses} FLOWER relies on a regularizer $\mathcal{R}_{PMAS}$ integrating the projecting concept \cite{Cha2021CPRCR} into the memory aware synapses approach to address the catastrophic forgetting problem. The application of projection concept induces wide local optimum regions improving scalability of a continual learner since it increases a chance of an overlapping region to be found. The PMAS method is formalized:  
\begin{equation}\label{projection}
        \begin{split}
        \mathcal{R}_{PMAS}=\frac{\lambda_{3}}{N_k}\sum_{i=1}^{N_k}\mathcal{L}_{KL}(g_{W_{C}}(f_{W_{F}}(x_i^k)),P_{U})\\
        +\lambda_{4}\sum_{i}\Xi_{i}^{k-1}(\Theta_{i}^{k}-\Theta_{i}^{k-1})^{2}
        \end{split}
\end{equation}
where $\Xi$ is a parameter importance matrix which can be estimated using different approaches: EWC \cite{kirkpatrick2016overcoming}, SI \cite{Zenke2017ContinualLT}, MAS \cite{Aljundi2018MemoryAS}, etc. We apply the MAS method based on the online and unsupervised method, since it is faster to compute than EWC and SI. $\Theta=\{W_{F},W_{C}\}$ where no regularization mechanism is applied to the transformation module $\kappa_{W_T}(.)$ because it only functions to enrich the current concept. The first term of \eqref{projection} promotes wide local minimum where $\mathcal{L}_{KL}(.)$ denotes the KL divergence loss function minimizing the discrepancies of the two components while the second term of \eqref{projection} is the MAS-based regularization approach inspired by the $L_2$ regularization technique except that of $(\Theta_i^{k}-\Theta_i^{k-1})$. $\Theta_i^{k-1},\Theta_i^{k}$ respectively denote the network parameters before and after seeing the current $k-th$ task. $\lambda_3,\lambda_4$ stand for predefined constants controlling the influence of each term. 

\eqref{projection} introduces the KL divergence term to induce wide local optimum regions where it is perceivable as the optimization $\min_{Q\in\mathcal{Q}}\mathcal{L}_{KL}(Q|P)$  where $P$ is a given distribution and $\mathcal{Q}$ is a convex set of distributions in the probability simplex $\Delta_m$ \cite{Cha2021CPRCR} where $g_{W_C}(.):\Re^{D}\rightarrow\Delta_{M}$ and $M$ is the label space dimension. After minimizing this term, $P^*$ is obtained and represents a distribution in $\mathcal{Q}$ being the closest to $P$. Such case presents $\mathcal{L}_{KL}(Q|P)$ as the Euclidean distance where $(Q, P^*, P)$ constructs a right triangle. 

\begin{figure}[!t]
 \centering
 \includegraphics[scale=0.5]{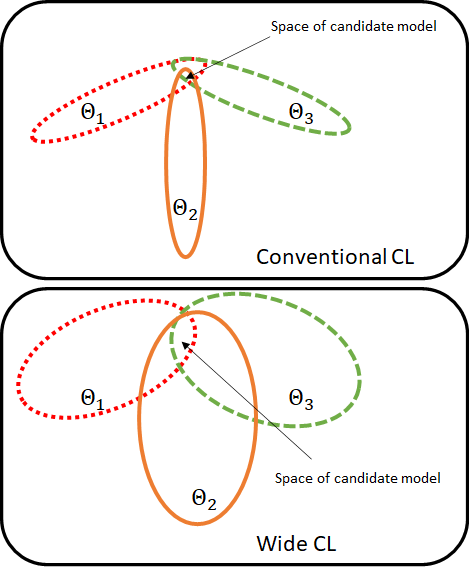}
 \caption{Wide Continual Learning: conventional continual learning approaches generate narrow ellipsoids where the local minima region is sharp leaving very small space for model candidate parameters, i.e., intersections of all ellipsoids. On the contrary, the wide continual learning generates wide ellipsoids having wide local minima regions allowing enough space of model candidate parameters.}
 \label{WideCL}
 \end{figure}

\noindent\textbf{Lemma 1} \cite{Cha2021CPRCR}. Suppose $P^*\in Q$ such that $\mathcal{L}_{KL}(P^*|P)=\min_{Q\in\mathcal{Q}}\mathcal{L}_{KL}(Q|P)$, then 
\begin{equation}
    \mathcal{L}_{KL}(Q|P)\geq \mathcal{L}_{KL}(Q|P^*)+\mathcal{L}_{KL}(P^*|P),\forall Q\in \mathcal{Q}
\end{equation}
The classifier's output $g_{W_C}(.)$ is projected to a uniform distribution in a set $\mathcal{C}$ and interpreted as conditional distributions $Q_{Y|X}$ between target $Y$ and input $X$. After learning the $k-th$ task with \eqref{projection}, $P^{k*}_{Y|X}\in\mathcal{C}$ is obtained. Suppose the presence of two classifiers in the $k+1-th$ task $P^{k+1}_{Y|X}\notin\mathcal{C}$ w/o the projection and $P^{(k+1)*}_{Y|X}\in\mathcal{C}$ with the projection, a right triangle is established across $(P^{k*}_{Y|X},P^{(k+1)}_{Y|X},P^{(k+1)*}_{Y|X})$ as per Lemma 1. $P^{k*}_{Y|X}$ is closer to $P^{(k+1)*}_{Y|X}$ than that without the projection $P^{(k+1)}_{Y|X}$ when being evaluated on the $k-th$ task. A uniform distribution $P_{U}$ is selected because it represents the centroid of $\Delta_{M}$ that results in an upper bound of divergence $\log{M}$. Note that an ideal classifier performing well on all tasks does not exist in the continual learning context and thus we turn our attention to the set of possible classifiers $\mathcal{C}$ to be a KL divergence ball centered at the uniform distribution $P_{U}$ \cite{Cha2021CPRCR}.  The wide learning strategy is illustrated in Fig. \ref{WideCL}.    

\subsubsection{Loss Function of Few-Shot Learning Phase} 
\begin{figure*}[!t]
 \centering
 \includegraphics[scale=0.45]{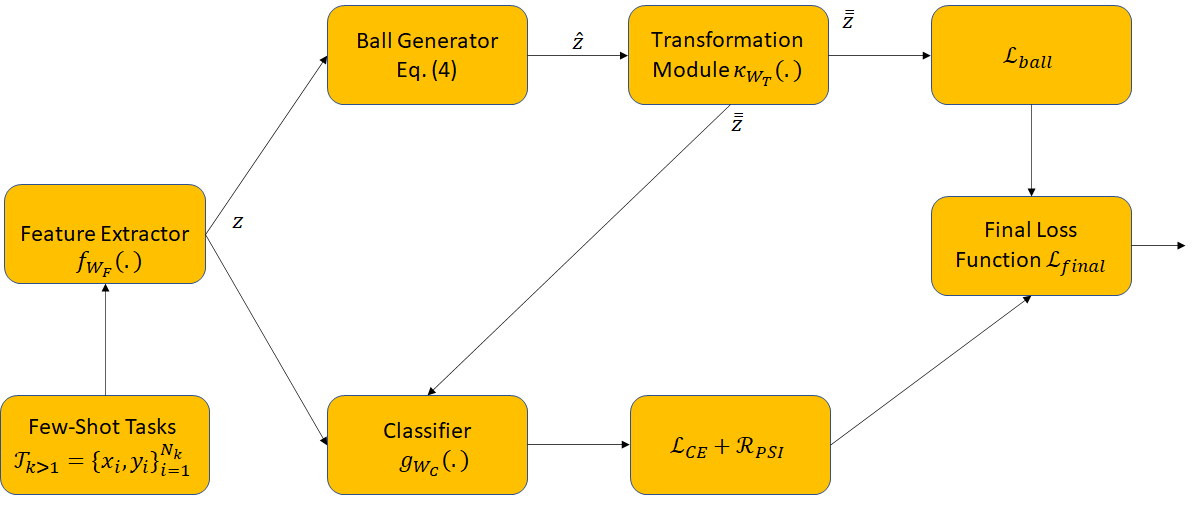}
 \caption{Few Shot Learning Phase of FLOWER is driven by the three loss functions, the cross entropy loss function, the ball generator loss function and the projection-based synaptic intelligence regularizer. It combines wide and flat learning paradigms supported by the feature space data augmentation module having a transformation module.}
 \label{fig_framework}
 \end{figure*}

The few-shot learning phase of FLOWER is controlled by three loss functions: cross entropy loss function $\mathcal{L}_{ce}$, ball generator loss function $\mathcal{L}_{ball}$ and regularizer $\mathcal{R}_{PMAS}$. The cross entropy loss function \eqref{LCE} learns the current concept supported by the feature space data augmentation method while the ball generator loss function \eqref{LBALL} addresses the synthetic sample bias problem where a synthetic sample is pulled closely to its centroid and pushed away from other centroids. The regularizer \eqref{projection} functions to protect against the catastrophic forgetting problem: 
\begin{equation}\label{LFINAL}
\mathcal{L}_{k>1}=\mathcal{L}_{ce}+\mathcal{L}_{ball}+\mathcal{R}_{PMAS}
\end{equation}
The network parameters are maintained in such a way to satisfy the b-flat-wide local minima conditions. That is, parameters of feature extractor are clamped if they go beyond the flat-wide region. Besides, the wide learning paradigm is actualized in the projection mechanism to scale FLOWER up to long sequences of tasks. \eqref{LFINAL} takes the advantage of the flat learning concept \cite{Shi2021OvercomingCF} and the wide learning concept \cite{Cha2021CPRCR} to address the CF problem without any memory and thus realizes the flat-wide learning strategy. Pseudo-codes of FLOWER are provided in the algorithm 1 while overview of flat-wide learning strategies is pictorially illustrated in Fig. \ref{fig_framework}.

\begin{algorithm}
\caption{Learning Policy of FLOWER}\label{algorithm1}
\begin{algorithmic}[1]
\State \textbf{Input:} The flat-wide region bound $b$, randomly initialized parameters $\Theta=\{W_{F},W_{C}\},W_{T}$, hyper-parameters $\lambda_{1},\lambda_{2},\lambda_{3},\lambda_{4}$, number of epochs $E$, number of synthetic samples $S$.
\State \textbf{Output:}  Updated Network Parameters $\Theta=\{W_{F},W_{C}\}$
\item[\hspace{5 mm} //Base Learning Phase $k=1$]
\For{$e=1:E$}
    \For{$j=1:M$} 
        \State Sample a noise vector $\epsilon_j\backsim\mathcal{E}$ where $-b\leq\epsilon_j\leq b$
        \State Perturb the feature extractor parameters $\Theta=\{W_{F}+\epsilon_j,W_{C}\}$
        \State Compute the base loss $\mathcal{L}_{base}$
        \State Reset the network parameters $\Theta=\{W_{F},W_{C}\}$
    \EndFor
    \State Update network parameters with $\mathcal{L}_{k=1}$ as per (M.3)
\EndFor

\item[\hspace{5 mm} //Few Shot Learning Phase $k>1$]
\For{$k=2:K$}
    \For{$e=1:E$}
        \State $\mathcal{T}'_{k}=\emptyset$
        \State Calculate the ball/class parameters $\omega(Z_c)=\{C_c,\sigma_c\}$
        \For{$i'=1:S$}
            \State Generate a synthetic sample $\hat{z}_{i'}$ as per (M.9)
            \State Apply the transformation module $\hat{\hat{z}}=\kappa_{W_T}(\hat{z}_{i'})$ to induce a 
            \item  [\hspace{20.5 mm} unbiased synthetic sample $\hat{\hat{z}}_{i'}$]
            \State $\mathcal{T}'_{k}=\mathcal{T}'_{k}\cup \{\hat{\hat{z}}_{i'},y_i\}$
        \EndFor
    \State $\hat{\mathcal{T}}_k^{e}=\mathcal{T}'_{k}\cup\mathcal{T}_k$
    \State Calculate the ball generator loss (M.10) with $\hat{\mathcal{T}}_{k}^{e}$
    \State Calculate the cross entropy loss (M.2) with $\hat{\mathcal{T}}_{k}^{e}$
    \State Calculate the final loss (M.13)
    \State Update network parameters $\Theta=\{W_{F},W_{C}\},W_T$ with $\mathcal{L}_{k}$
    \State Clamp feature extractor parameters $W_{F^*}-b\leq{W}_{F}\leq W_{F^*}+b$ to
    \item  [\hspace{15.5 mm} meet the b-flat-wide local minima conditions]
    \EndFor
\EndFor
\end{algorithmic}
\end{algorithm}

\section{Experiments}
This section presents numerical validations of FLOWER: comparisons with prior art approaches with different sizes of base tasks, ablation study and sensitivity analyses of the bound $b$ and the number of trials $M$.

\subsection{Experimental Setting}
\noindent\textbf{Datasets:} our numerical study is carried out using CIFAR100, miniImageNet and CUB-200-2011. The CIFAR100 and miniImageNet problems are configured such that 60 classes are reserved for the base task while the remaining 40 classes as the few-shot tasks under the 5-way 5-shots configuration, i.e., 5 classes are sampled where each class comprises 5 samples, thus leading to 8 few-shot tasks. For the CUB-200-2011 problem with 200 classes, the base task is built upon 100 classes while the few-shot tasks utilizes the remainder 100 classes with the 10-way 5-shot configuration, thus forming 10 few-shot tasks. In addition, we also evaluate consolidated algorithms in the case of small base tasks where only 20 classes are included in the miniimagenet and CIFAR100 problems while 50 classes are presented for the CUB problem. Numerical results under a 1-shot configuration are also reported here. 

\noindent\textbf{Benchmark Algorithms:} FLOWER is compared with 7 state-of-the-art algorithms: iCaRL \cite{Rebuffi2017iCaRLIC}, Rebalance \cite{Hou2019LearningAU}, FSLL \cite{Mazumder2021FewShotLL}, cRT \cite{Kang2020DecouplingRA}, TOPIC \cite{Tao2020FewShotCL}, EEIL \cite{Castro2018EndtoEndIL} and F2M \cite{Shi2021OvercomingCF}. The baseline algorithm follows \cite{Shi2021OvercomingCF} where the training process is only carried out in the base task while only exemplar constructions are done for few-shot tasks for a simple nearest class mean classification algorithm. A joint training is also included where the training process encompasses base classes and few-shot classes. Although the joint-training method commonly serves as \textbf{the upper bound} for the continual learning problems, it is not strong enough here because the extreme imbalance problem between the base task and the few-shot tasks exists in the few-shot continual learning problems. Here, the cRT is added as \textbf{the additional upper bound} because it performs long-tailed classification trained with all encountered data \cite{Shi2021OvercomingCF}. Numerical results of ICARL, Rebalance, EEIL, TOPIC, cRT and joint-training are taken from \cite{Tao2020FewShotCL} and \cite{Shi2021OvercomingCF} while we run ICARL, Rebalance Baseline, F2M and FSLL in the same computational environments applying the implementation of \cite{Shi2021OvercomingCF} with their best hyper-parameters found with the grid search.

\noindent\textbf{Experimental Details:} our numerical study is executed under single NVIDIA A100 GPU with 40 GB memory across five runs. ResNet18 is adopted as the backbone network for all methods in the miniimagenet and CUB-200-2011 problems, while ResNet20 is implemented as a backbone network for all methods in the CIFAR100 problem. As with \cite{Shi2021OvercomingCF}, $M$ is set in between 2 to 4 and noise is injected to the last 4 or 8 convolutional layers while the flat region bound $b$ is assigned as 0.01. Epoch per session is selected as 15 and the learning rate is assigned as 0.2 initially and truncated by 1e-6 after 5 epochs. Hyper-parameters of all methods are detailed in Table \ref{tab:hyperparameters}.

\begin{table}[]
\centering
\begin{tabular}{ll}
\hline
{\textbf{hyperparameter}}                               & {\textbf{value}}                         \\ \hline
{image\_size}                                           & {32 (CIFAR), 84 (MiniImageNet), 224(CUB)}\\ \hline
{networks}                                              & {ResNet18, ResNet20(CIFAR)}              \\ \hline
{optimizer}                                             & {SGD}                                    \\ \hline
{bacth\_size}                                           & {256}                                    \\ \hline
{\begin{tabular}[c]{@{}l@{}}epoch\\ - base task\\ - continual tasks\end{tabular}}                                 & {\begin{tabular}[c]{@{}l@{}} \\ - 1000, 600 (CUB)\\ - 15, generate prototype only (CUB)\end{tabular}}  \\ \hline
{\begin{tabular}[c]{@{}l@{}}initial learning rate (lr)\\ - base task\\ - continual tasks\end{tabular}}             & {\begin{tabular}[c]{@{}l@{}}\\ - 0.1, 0.001 (CUB)\\ - 0.2\end{tabular}}                               \\ \hline
{\begin{tabular}[c]{@{}l@{}}gamma\\ - base task\\ - continual tasks\end{tabular}}                                  & {\begin{tabular}[c]{@{}l@{}} \\ - 0.1, 1e-6 (CUB)\\ - 1e-6\end{tabular}}                               \\ \hline
{weight\_decay}                                         & {5e-4}                               \\ \hline
{momentum}                                              & {0.9}                                    \\ \hline
{\begin{tabular}[c]{@{}l@{}}random\_noise\\ - distribution\\ - type\\ - num\_layers\\ - low\\ - high\\ - reduction\_factor\\ - bound\_value\\ - random\_times\end{tabular}} & {\begin{tabular}[c]{@{}l@{}} \\ - DiscreteBeta\\ - suffix\_conv\_weight\\ - 4 or 8\\ - 0.1\\ - 5.0, 3.0 (CUB)\\ - 4\\ - 0.01, 0.0002 (SUB)\\ - 2-4\end{tabular}} \\ \hline
{\begin{tabular}[c]{@{}l@{}}lambda\\ - lambda1\\ - lambda2\\ - lambda3\\ - lambda4\end{tabular}}                  & {\begin{tabular}[c]{@{}l@{}} \\ - 1.0\\ - 1.0\\ - 10\\ - 100\end{tabular}}    \\ \hline                        
\end{tabular}
\caption{Hyper-parameters setting for our experiments}
\label{tab:hyperparameters}
\end{table}

\begin{table*}
\small
\centering
\begin{tabular}{lccccccccccc}
\hline
\cline{1-12}
{}                                                           & \multicolumn{9}{c}{{\textbf{Session}}}         & {}                               & {}                          \\ \cline{2-10}
\multirow{-2}{*}{{\textbf{Method}}} & \centering{1}              & \centering{2}              & \centering{3}              & \centering{4}              & \centering{5}              & \centering{6}              & \centering{7}              & \centering{8}              & \centering{9}              & \multirow{-2}{*}{{\textbf{Avg}}} & \multirow{-2}{*}{{\textbf{Gap}}} \\ \cline{1-12} \hline 
\multicolumn{12}{c}{{\textbf{Base classes = 60}}} \\ \hline
{cRT* \cite{Kang2020DecouplingRA}}                                                         & {67.30}          & {64.15}          & {60.59}          & {57.32}          & {54.22}          & {51.43}          & {48.92}          & {46.78}          & {44.85}          & {55.06}                          & {0.94}                      \\
{\begin{tabular}[]{@{}l@{}}Joint-\\ training* \cite{Shi2021OvercomingCF}\end{tabular}}                                              & {67.30}          & {62.34}          & {57.79}          & {54.08}          & {50.93}          & {47.65}          & {44.65}          & {42.61}          & {40.29}          & {51.96}                          & {-2.17}                     \\
{Baseline \cite{Shi2021OvercomingCF}}                                                       & {66.78}          & {61.29}          & {57.25}          & {54.17}          & {51.54}          & {49.00}          & {46.56}          & {44.99}          & {{43.78}} & {52.82}                          & {-1.31}                     \\ [0.6ex] 
{F2M \cite{Shi2021OvercomingCF}}                                                        & {66.07}          & {61.05}          & {56.82}          & {53.51}          & {50.76}          & {48.26}          & {45.79}          & {44.07}          & {42.62}          & {52.11}                          & {-2.02}                     \\ [0.6ex] 
{FSLL \cite{Mazumder2021FewShotLL}}                                                        & {66.98}          & {57.23}          & {52.47}          & {49.66}          & {47.45}          & {45.20}          & {43.29}          & {42.06}          & {41.18}          & {49.50}                          & {-4.62}                     \\ [0.6ex] 
{ICARL \cite{Rebuffi2017iCaRLIC}}                                                    & {66.05}          & {56.47}          & {53.26}          & {50.14}          & {47.55}          & {45.08}          & {42.47}          & {41.04}          & {39.60}          & {49.07}                          & {-5.05}                     \\ [0.6ex]
{Rebalance \cite{Hou2019LearningAU}}                                        & {66.45}          & {60.66}          & {55.61}          & {51.38}          & {47.93}          & {44.64}          & {41.40}          & {38.75}          & {37.11}          & {49.33}                          & {-4.80}                     \\ [0.6ex]
{ICARL+ \cite{Rebuffi2017iCaRLIC}}                                               & {61.31}          & {46.32}          & {42.94}          & {37.63}          & {30.49}          & {24.00}          & {20.89}          & {18.80}          & {17.21}          & {33.29}                          & {-20.84}                    \\ [0.6ex]
{EEIL+ \cite{Castro2018EndtoEndIL}}                              & {61.31}          & {46.58}          & {44.00}          & {37.29}          & {33.14}          & {27.12}          & {24.10}          & {21.57}          & {19.58}          & {34.97}                          & {-19.16}                    \\ [0.6ex]
{Rebalance+ \cite{Hou2019LearningAU}}                                              & {61.31}          & {47.80}          & {39.31}          & {31.91}          & {25.68}          & {21.35}          & {18.67}          & {17.24}          & {14.17}          & {30.83}                          & {-23.30}                    \\ [0.6ex]
{TOPIC+ \cite{Tao2020FewShotCL}}                                                   & {61.31}          & {50.09}          & {45.17}          & {41.16}          & {37.48}          & {35.52}          & {32.19}          & {29.46}          & {24.42}          & {39.64}                          & {-14.48}                    \\ [0.6ex]
{\textbf{FLOWER}}                   & {\textbf{68.83}} & {\textbf{63.27}} & {\textbf{59.00}} & {\textbf{55.61}} & {\textbf{52.64}} & {\textbf{49.96}} & {\textbf{47.56}} & {\textbf{45.86}} & {\textbf{44.40}} & {\textbf{54.13}}                 & {\textbf{0.00}}             \\ [0.6ex] \hline
\multicolumn{12}{c}{{\textbf{Base classes = 20}}} \\ \hline
{Baseline \cite{Shi2021OvercomingCF}}                                        & {67.90}          & {46.32}          & {37.24}          & {30.13}          & {25.56}          & {22.11}          & {20.03}          & {18.14}          & {16.80}          & {31.58}                          & {-3.65}                     \\ [0.6ex]
{F2M \cite{Shi2021OvercomingCF}}                                                          & {65.45}          & {45.70}          & {36.15}          & {29.19}          & {24.73}          & {21.29}          & {19.18}          & {17.18}          & {15.91}          & {30.53}                          & {-4.70}                     \\ [0.6ex]
{FSLL \cite{Mazumder2021FewShotLL}}                                                       & {68.05}          & {47.05}          & {37.74}          & {30.70}          & {26.00}          & {22.43}          & {20.26}          & {18.22}          & {17.11}          & {31.95}                          & {-3.28}                     \\ [0.6ex]
{ICARL \cite{Rebuffi2017iCaRLIC}}                                                      & {60.70}          & {38.91}          & {30.65}          & {25.30}          & {21.46}          & {17.97}          & {16.38}          & {14.58}          & {13.77}          & {26.64}                          & {-8.59}                     \\ [0.6ex]
{Rebalance \cite{Hou2019LearningAU}}                                              & {67.55}          & {44.89}          & {33.67}          & {29.04}          & {25.32}          & {22.19}          & {20.73}          & {18.88}          & {17.70}          & {31.11}                          & {-4.12}                     \\ [0.6ex]
{\textbf{FLOWER}}                         & {\textbf{75.70}} & {\textbf{52.66}} & {\textbf{41.67}} & {\textbf{33.88}} & {\textbf{28.64}} & {\textbf{24.63}} & {\textbf{22.00}} & {\textbf{19.67}} & {\textbf{18.21}} & {\textbf{35.23}}                 & {\textbf{0.00}}  \\ [0.6ex] \hline
\end{tabular}
\caption{Classification accuracy on split MiniImagenet dataset with 60 and 20 base classes averaged across 5 times run, \#B indicates the number of base classes, * indicates the results are copied from F2M, + indicates the results copied from TOPIC.}
\label{tab:miniimagenet}
\end{table*}

\begin{table*}[]
\small
\centering
\begin{tabular}{lccccccccccc}
\hline
\cline{1-12}
{}       & \multicolumn{9}{c}{{\textbf{Session}}}        & {}    & {} \\ \cline{3-11} 
\multirow{-2}{*}{{\textbf{Method}}} & \centering{1}              & \centering{2}              & \centering{3}              & \centering{4}              & \centering{5}              & \centering{6}              & \centering{7}              & \centering{8}              & \centering{9}   & \multirow{-2}{*}{{\textbf{Avg}}} & \multirow{-2}{*}{{\textbf{Gap}}}\\ \cline{1-12} \hline
\multicolumn{12}{c}{{\textbf{Base classes = 60}}} \\ \hline
{cRT* \cite{Kang2020DecouplingRA}}   & {65.18}         & {63.89}      & {60.20}          & {57.23}          & {53.71}          & {50.39}          & {48.77}          & {47.29}          & {45.28}          & {54.66}  & {-4.76}        \\
{\begin{tabular}[c]{@{}l@{}}Joint-\\ training* \cite{Shi2021OvercomingCF}\end{tabular}}        & {65.18}         & {61.45}          & {57.36}          & {53.68}          & {50.84}          & {47.33}          & {44.79}          & {42.62}          & {40.08}          & {51.48}  & {-7.94}        \\
{Baseline \cite{Shi2021OvercomingCF}}   & {72.2}          & {67.88}          & {64.06}          & {60.43}          & {57.28}          & {54.54}          & {52.83}          & {50.93}          & {\textbf{48.8}}  & {58.77}  & {-0.65}        \\ [0.6ex]
{F2M \cite{Shi2021OvercomingCF}}    & {71.4}          & {66.65}          & {63.2}           & {59.54}          & {56.61}          & {54.08}          & {52.2}           & {50.49}          & {48.4}           & {58.06}  & {-1.36}        \\ [0.6ex]
{FSLL \cite{Mazumder2021FewShotLL}}    & {72.52}         & {64.2}           & {59.6}           & {55.09}          & {52.85}          & {51.57}          & {51.21}          & {49.66}          & {47.86}          & {56.06}  & {-3.36}        \\ [0.6ex]
{ICARL \cite{Rebuffi2017iCaRLIC}}   & {71.72}         & {64.86}          & {60.46}          & {56.61}          & {53.76}          & {51.1}           & {49.1}           & {46.95}          & {44.64}          & {55.47}  & {-3.95}        \\ [0.6ex]
{Rebalance \cite{Hou2019LearningAU}}    & {74.57}         & {66.65}          & {60.96}          & {55.59}          & {50.87}          & {46.55}          & {43.82}          & {40.29}          & {37.23}          & {52.95}  & {-6.47}        \\ [0.6ex]
{ICARL+ \cite{Rebuffi2017iCaRLIC}}   & {64.10}         & {53.28}          & {41.69}          & {34.13}          & {27.93}          & {26.06}          & {20.41}          & {15.48}          & {13.73}          & {32.98}  & {-26.44}       \\ [0.6ex]
{EEIL+ \cite{Castro2018EndtoEndIL}}   & {64.10}         & {53.11}          & {42.71}          & {35.15}          & {28.96}          & {24.98}          & {21.01}          & {17.26}          & {15.85}          & {33.68}  & {-25.74}       \\ [0.6ex]
{Rebalance+ \cite{Hou2019LearningAU}}     & {64.10}         & {53.05}          & {43.96}          & {36.97}          & {31.61}          & {26.73}          & {21.23}          & {16.78}          & {13.54}          & {34.22}  & {-25.20}       \\ [0.6ex]
{TOPIC+ \cite{Tao2020FewShotCL}}    & {64.10}         & {55.88}          & {47.07}          & {45.16}          & {40.11}          & {36.38}          & {33.96}          & {31.55}          & {29.37}          & {42.62}  & {-16.80}       \\ [0.6ex]
{\textbf{FLOWER}}         & {\textbf{73.4}} & {\textbf{68.98}} & {\textbf{64.98}} & {\textbf{61.2}}  & {\textbf{57.88}} & {\textbf{55.14}} & {\textbf{53.28}} & {\textbf{51.16}} & {\textbf{48.78}} & {\textbf{59.42}}     & {\textbf{0.00}}            \\ [0.6ex]   \hline
\multicolumn{12}{c}{{\textbf{Base classes = 20}}} \\ \hline
{Baseline \cite{Shi2021OvercomingCF}}   & {74.75}         & {54.39}          & {41.83}          & {35.24}          & {30.72}          & {27.67}          & {24.91}          & {22.82}          & {20.94}          & {37.03}  & {-3.29}        \\ [0.6ex]
{F2M \cite{Shi2021OvercomingCF}}   & {73.4}          & {53.25}          & {42.1}           & {35.42}          & {31.03}          & {28.12}          & {24.95}          & {23.04}          & {21.38}          & {36.97}  & {-3.35}        \\ [0.6ex]
{FSLL \cite{Mazumder2021FewShotLL}}   & {75}            & {52.76}          & {39.87}          & {33.29}          & {29.74}          & {26.73}          & {24.12}          & {22.22}          & {20.28}          & {36.00}  & {-4.32}        \\ [0.6ex]
{ICARL \cite{Rebuffi2017iCaRLIC}}   & {76.85}         & {53.58}          & {41.51}          & {34.65}          & {30.22}          & {27.15}          & {23.73}          & {21.65}          & {19.62}          & {36.55}  & {-3.77}        \\ [0.6ex]
{Rebalance \cite{Hou2019LearningAU}}    & {73.9}          & {49.27}          & {36.95}          & {31.73}          & {28.62}          & {26.68}          & {24.44}          & {22.62}          & {20.86}          & {35.01}  & {-5.31}        \\ [0.6ex]
{\textbf{FLOWER}}            & {\textbf{83.2}} & {\textbf{58.93}} & {\textbf{46.01}} & {\textbf{38.56}} & {\textbf{33.48}} & {\textbf{29.54}} & {\textbf{26.49}} & {\textbf{24.4}}  & {\textbf{22.27}} & {\textbf{40.32}}     & {\textbf{0.00}}  \\ [0.6ex]  \hline       
\end{tabular}
\caption{Classification accuracy on split CIFAR100 dataset with 60 and 20 base classes averaged across 5 times run, \#B indicates the number of base classes, * indicates the results are copied from F2M, + indicates the results copied from TOPIC.}
\label{tab:cifar100}
\end{table*}

\begin{table*}[]
\small
\centering
\begin{tabular}{lccccccccccccc}
\hline
\cline{1-14}
{}       & \multicolumn{11}{c}{{\textbf{Session}}}& {}    & {}  \\ \cline{2-12}
\multirow{-2}{*}{{\textbf{Method}}}  & \centering{1}              & \centering{2}              & \centering{3}              & \centering{4}              & \centering{5}              & \centering{6}              & \centering{7}              & \centering{8}              & \centering{9} & \centering{10} & \centering{11} & \multirow{-2}{*}{{\textbf{Avg}}} & \multirow{-2}{*}{{\textbf{Gap}}} \\ \cline{1-14}  \hline 
\multicolumn{14}{c}{{\textbf{Base classes = 100}}} \\ \hline
\text{cRT*\cite{Kang2020DecouplingRA}}   & \text{80.83}          & \text{78.51}          & \text{76.12}          & \text{73.93}          & \text{71.46}          & \text{68.96}          & \text{67.73}          & {66.75}          & {64.22}          & {62.53}          & {61.08}          & {70.19} & {4.52}            \\
\text{\begin{tabular}[c]{@{}l@{}}Joint-\\ training* \cite{Shi2021OvercomingCF}\end{tabular}}      & {80.83}          & {77.57}          & {74.11}          & {70.75}          & {68.52}          & {65.97}          & {64.58}          & {62.22}          & {60.18}          & {58.49}          & {56.78}          & {67.27}& {1.60}            \\
\text{Baseline \cite{Shi2021OvercomingCF}}    & {76.92}          & {72.83}          & {68.96}          & {64.74}          & {62.38}          & {59.63}          & {58.44}          & {57.25}          & {{54.97}} & {54.33}          & {53.49}          & {62.18}& {-3.49}           \\ [0.6ex] 
\text{F2M \cite{Shi2021OvercomingCF}}     & {77.41}          & {73.5}           & {69.52}          & {65.27}          & {63.07}          & {60.41}          & {59.2}           & {58.02}          & {55.89}          & {55.49}          & {54.51}          & {62.94}& {-2.73}           \\ [0.6ex] 
\text{FSLL \cite{Mazumder2021FewShotLL}}     & {76.92}          & {70.58}          & {64.73}          & {57.77}          & {55.96}          & {54.34}          & {53.62}          & {53.04}          & {50.45}          & {50.36}          & {49.48}          & {57.93}& {-7.74}           \\ [0.6ex] 
\text{ICARL$\circ$ \cite{Rebuffi2017iCaRLIC}}      & {75.73}          & {69.97}          & {65.95}          & {61.27}          & {57.62}          & {55.5}           & {54.16}          & {53.37}          & {50.75}          & {50.35}          & {48.77}          & {58.49}& {-7.18}           \\ [0.6ex] 
\text{Rebalance \cite{Hou2019LearningAU}}            & {74.83}          & {55.51}          & {48.36}          & {42.38}          & {36.98}          & {32.82}          & {30.15}          & {27.82}          & {25.96}          & {24.44}          & {22.83}          & {38.37}& {-27.30}          \\ [0.6ex] 
\text{ICARL+ \cite{Rebuffi2017iCaRLIC}}    & {68.68}          & {52.65}          & {48.61}          & {44.16}          & {36.62}          & {29.52}          & {27.83}          & {26.26}          & {24.01}          & {23.89}          & {21.16}          & {36.67}& {-29.00}          \\ [0.6ex] 
\text{EEIL+ \cite{Castro2018EndtoEndIL}}    & {68.68}          & {53.63}          & {47.91}          & {44.20}          & {36.30}          & {27.46}          & {25.93}          & {24.70}          & {23.95}          & {24.13}          & {22.11}          & {36.27}& {-29.40}          \\ [0.6ex] 
\text{Rebalance+ \cite{Hou2019LearningAU}}        & {68.68}          & {62.55}          & {50.33}          & {45.07}          & {38.25}          & {32.58}          & {28.71}          & {26.28}          & {23.80}          & {19.91}          & {17.82}          & {37.63}& {-28.04}          \\ [0.6ex] 
\text{TOPIC+ \cite{Tao2020FewShotCL}}   & {68.68}          & {62.49}          & {54.81}          & {49.99}          & {42.25}          & {41.40}          & {38.35}          & {35.36}          & {32.22}          & {28.31}          & {26.28}          & {43.65}& {-22.02}          \\ [0.6ex] 
{\textbf{FLOWER}}& {\textbf{79.02}} & {\textbf{75.77}} & {\textbf{72.01}} & {\textbf{67.96}} & {\textbf{65.99}} & {\textbf{63.38}} & {\textbf{62.14}} & {\textbf{61.12}} & {\textbf{58.96}} & {\textbf{58.52}} & {\textbf{57.49}} & {\textbf{65.67}}    &  {\textbf{0.00}} \\  \hline 
\multicolumn{14}{c}{{\textbf{Base classes = 50}}} \\ \hline
\text{Baseline \cite{Shi2021OvercomingCF}}  & {77.83}          & {66.94}          & {61.18}          & {59.13}          & {54.85}          & {49.96}          & {46.73}          & {43.93}          & {42.3}           & {39.98}          & {39.74}          & {52.96}& {-4.86}           \\ [0.6ex] 
\text{F2M \cite{Shi2021OvercomingCF}}    & {78.11}          & {68.52}          & {63.5}           & {61.61}          & {57.58}          & {52.64}          & {49.26}          & {46.43}          & {44.81}          & {42.71}          & {42.41}          & {55.23}& {-2.59}           \\ [0.6ex] 
\text{FSLL \cite{Mazumder2021FewShotLL}}      & {77.61}          & {65.96}          & {60.37}          & {58.59}          & {55.96}          & {51.75}          & {48.35}          & {45.84}          & {44.92}          & {42.8}           & {42.62}          & {54.07}& {-3.75}           \\ [0.6ex] 
\text{ICARL$\circ$ \cite{Rebuffi2017iCaRLIC}}  & {75.88}          & {62.54}          & {57.7}           & {54.73}          & {49.16}          & {43.96}          & {40.84}          & {38.09}          & {36.63}          & {34.38}          & {34.79}          & {48.06}& {-9.76}           \\ [0.6ex] 
\text{Rebalance \cite{Hou2019LearningAU}}    & {74.73}          & {60.91}          & {56.29}          & {53.93}          & {48.31}          & {42.49}          & {37.4}           & {33.42}          & {31.32}          & {29.29}          & {27.93}          & {45.09}& {-12.73}          \\ [0.6ex] 
{\textbf{FLOWER}}          & {\textbf{79.63}} & {\textbf{71.14}} & {\textbf{66.18}} & {\textbf{64.22}} & {\textbf{60.03}} & {\textbf{55.17}} & {\textbf{51.74}} & {\textbf{49.13}} & {\textbf{47.79}} & {\textbf{45.68}} & {\textbf{45.29}} & {\textbf{57.82}}    & {\textbf{0.00}}    \\  \hline        
\end{tabular}
\caption{Classification accuracy on split CUB 2011 dataset with 100 and 50 base classes averaged across 5 times run, \#B indicates the number of base classes, * indicates the results are copied from F2M, + indicates the results copied from TOPIC, $\circ$ indicates the method is run 1 time due to system crash.}
\label{tab:cub}
\end{table*}

\begin{table}[]
\small
\centering
\begin{tabular}{ccccccc}
\hline
{}                              & {}                                & {}                                & \multicolumn{3}{c}{{\textbf{Session}}}    & {}                               \\ \cline{4-6}
\multirow{-2}{*}{{\textbf{FM}}} & \multirow{-2}{*}{{\textbf{PMAS}}} & \multirow{-2}{*}{{\textbf{Ball}}} & {{1}}     & {{9}}     & {\textbf{1-9(Avg)}} & \multirow{-2}{*}{{\textbf{Gap}}} \\ \hline
{}                              & {\checkmark}                               & {\checkmark}                               & {67.82}          & {44.21}          & {53.77}             & {-0.36}                          \\
{\checkmark}                             & {}                                & {\checkmark}                               & {67.43}          & {44.20}          & {53.63}             & {-0.50}                          \\
{\checkmark}                             & {\checkmark}                               & {}                                & {68.83}          & {3.48}           & {40.43}             & {-13.70}                         \\
{\checkmark}                             & {\checkmark}                               & {\checkmark}                               & {\textbf{68.83}} & {\textbf{44.40}} & {\textbf{54.13}}    & {\textbf{0.00}} \\ \hline             
\end{tabular}
\caption{Ablation study on split MiniImagenet dataset with 60 base classes averaged across 5 times run.}
\label{tab:ablation}
\end{table}

\begin{table}[]
\small
\centering

\begin{tabular}{lccc}
\hline
{}                                  & \multicolumn{3}{c}{\textbf{Session}}                                                                    \\ \cline{2-4}
\multirow{-2}{*}{\textbf{Bound value (b)}} & {\textbf{1}}              & {\textbf{9}}              & {\textbf{1-9(Avg)}}       \\ \hline
{0.0001}                            & {67.03}          & {43.93}          & {53.26}          \\
{0.001}                             & {67.43}          & {43.53}          & {53.14}          \\
{0.0025}                            & {\textbf{68.43}} & {\textbf{44.68}} & {\textbf{54.48}} \\
{0.005}                             & {\textbf{68.37}} & {\textbf{44.77}} & {\textbf{54.29}} \\
{0.01}                              & {\textbf{68.83}} & {\textbf{44.40}} & {\textbf{54.13}} \\
{0.025}                             & {65.47}          & {41.73}          & {51.51}          \\
{0.05}                              & {65.93}          & {42.48}          & {52.05}          \\
{0.075}                             & {65.33}          & {42.50}          & {51.80}          \\
{0.1}                               & {64.45}          & {41.43}          & {50.85}  \\ \hline
\end{tabular}
\caption{Sensitivity analysis the impact of bound value to classification accuracy on MiniImageNet dataset with 60 base classes averaged across 5 time runs.}
\label{tab:bound value}
\end{table}

\begin{table}[]
\small
\centering
\begin{tabular}{cccc}
\hline
{}                                  & \multicolumn{3}{c}{\textbf{Session}}                                                                    \\ \cline{2-4}
\multirow{-2}{*}{\textbf{Random times (M)}} & {\textbf{1}}              & {\textbf{9}}              & {\textbf{1-9(Avg)}}       \\ \hline
{1*}                                  & {1.67}           & {-}              & {-}              \\
{2}                                  & {68.83}          & {44.40}          & {54.13}          \\
{3}                                  & {{68.75}} & {\textbf{44.65}} & {\textbf{54.52}} \\
{4}                                  & {\textbf{69.12}} & {\textbf{44.49}} & {\textbf{54.49}} \\
{5}                                  & {{68.73}} & {{43.74}} & {{53.89}} \\
{6}                                  & {68.45}          & {40.64}          & {53.26}          \\
{7}                                  & {68.77}          & {42.69}          & {53.57}          \\
{8}                                  & {\textbf{69.05}} & {\textbf{44.48}} & {\textbf{54.36}} \\
{9}                                  & {\textbf{69.13}} & {33.77}          & {50.12}          \\
{10}                                 & {68.45}          & {44.26}          & {54.04} \\ \hline   
\end{tabular}
\caption{Sensitivity analysis the impact of random times to classification accuracy on MiniImageNet dataset with 60 base classes averaged across 5 time runs, * denotes only run in base task due to poor performance}
\label{tab:random_times}
\end{table}

\begin{table*}[]
\footnotesize
\centering
\begin{tabular}{lcccccc}
\hline
{Dataset} & {Baseline\cite{Shi2021OvercomingCF}} & {F2M\cite{Shi2021OvercomingCF}}   & {FSLL\cite{Mazumder2021FewShotLL}}  & {ICARL\cite{Rebuffi2017iCaRLIC}} & {Rebalance\cite{Hou2019LearningAU}} & {\textbf{FLOWER}} \\ \hline
{MiniImageNet} & {51.59}                                                & {50.90} & {46.49} & {8.44}  & {51.19}   & {\textbf{53.11}} \\
{CIFAR-100}                                                 & {57.01}                                                & {56.07} & {52.89} & {9.07}  & {\textbf{57.44}}                                                 & {\textbf{57.43}}                                     \\
{CUB-2011}                                                 & {57.76}                                                & {57.73} & {52.86} & {8.22*} & {40.62}                                                 & {\textbf{59.27}}        \\ \hline                             
\end{tabular}
\caption{All sessions average accuracy with 1-shot setting on MiniImageNet, CIFAR-100 and CUB-200 dataset averaged across 5 time runs,* denotes only run 1 time due to system crash. \textit{Complete results are offered in the appendices}.}
\label{tab:1shot}
\end{table*}

\subsection{Numerical Results} 
Numerical results of consolidated algorithms across the CIFAR100, miniimagenet and CUB problems under moderate base tasks, e.g., 50 for CIFAR100 and miniimagenet and 100 for CUB, are reported in Tables \ref{tab:miniimagenet}, \ref{tab:cifar100}, \ref{tab:cub}, while the detailed (with standard deviation) results are tabulated in Table \ref{tab:full_miniimagenet}, \ref{tab:full_cifar100}, \ref{tab:full_cub}. FLOWER delivers the most encouraging performance in the CIFAR100 problem and even demonstrates better performances than the upper bounds, cRT and joint-training. It respectively beats baseline with $0.65\%$ gap and F2M with $1.36\%$ gap. The same finding is observed in realm of per session accuracy where FLOWER outperforms other consolidated algorithms. FLOWER consistently exceeds other algorithms in the miniimagenet problem even with higher differences than that of the CIFAR100 problem. It respectively beats the baseline with $1.31\%$ gap and the F2M with $2.02\%$ gap. FLOWER also delivers higher per session accuracy than those other algorithms. As with other two problems, FLOWER is superior to others in the CUB problem where it outperforms F2M ($2.73\%$ margin) and Baseline ($3.49\%$ margin) while consistently attains higher per session accuracy than other algorithms.  

FLOWER's performances are even more promising in the small base task than in the moderate base task where numerical results of consolidated algorithms are reported in Tables \ref{tab:miniimagenet}, \ref{tab:cifar100} and \ref{tab:cub}. It outperforms Baseline with $3.29\%$ gap and F2M with $3.35\%$ in the CIFAR100 problem. It achieves higher accuracy than FSLL in the miniimagenet problem with $3.28\%$ margin and F2M in the CUB problem with $2.59\%$ margin. The same pattern is observed for the context of per session accuracy. These support the efficacy of FLOWER for few-shot continual learning problems notably for small base tasks. We also confirm that the Baseline where only prototype constructions are performed in the few-shot learning sessions while leaving other parameters fixed is capable of producing strong performances as observed in \cite{Shi2021OvercomingCF}. That is, a model is supposed to retain decent performances of the base task here for the few-shot continual learning problems due to a large number of classes. 

\begin{figure}
\includegraphics[scale=0.37]{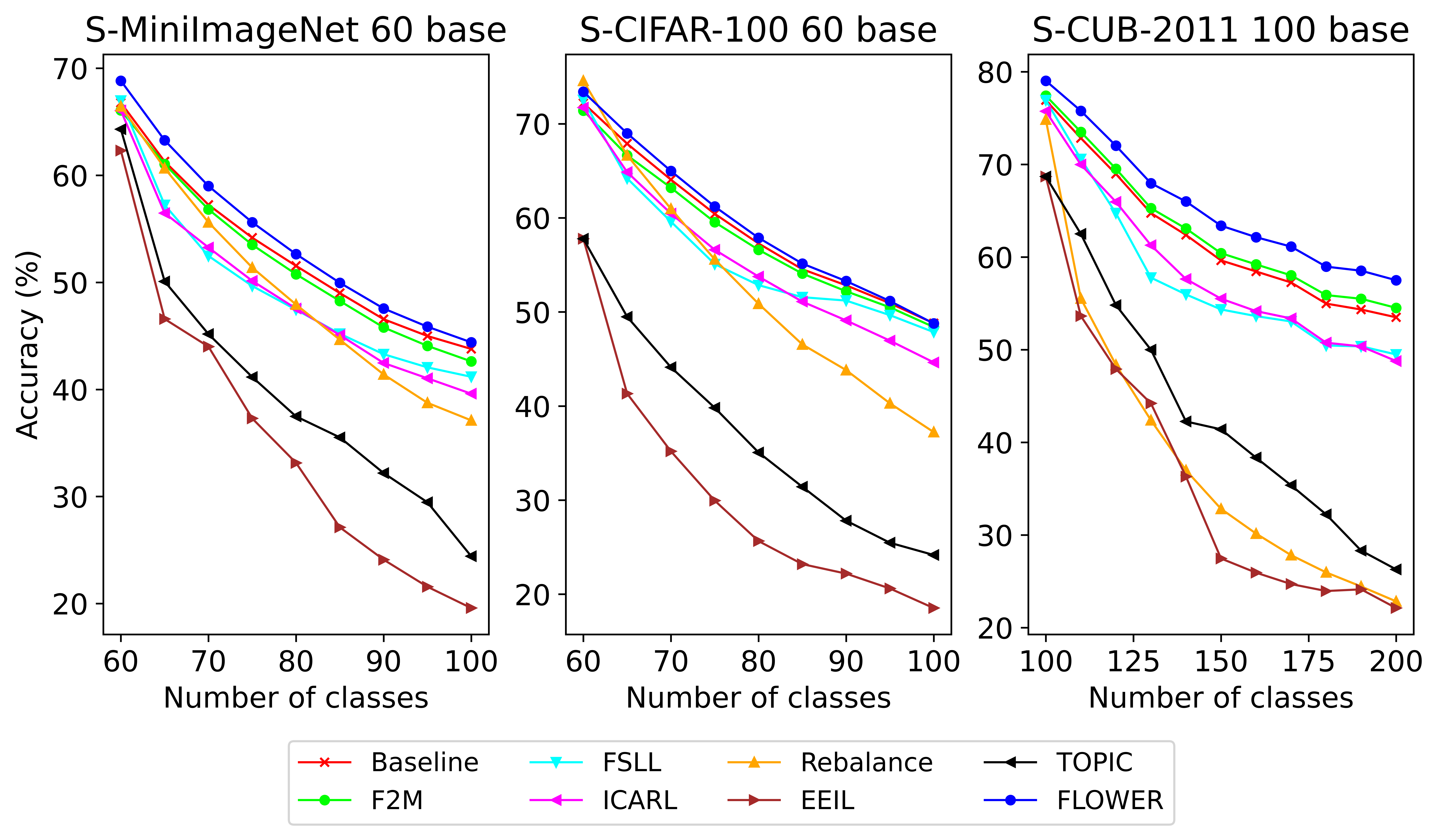}
\caption{Visualization of Performances of Consolidated Algorithms across three problems under the moderate base tasks}
\label{Fig:performance_plot}
\end{figure}

\begin{figure}[!t]
 \centering
 \includegraphics[scale=0.28]{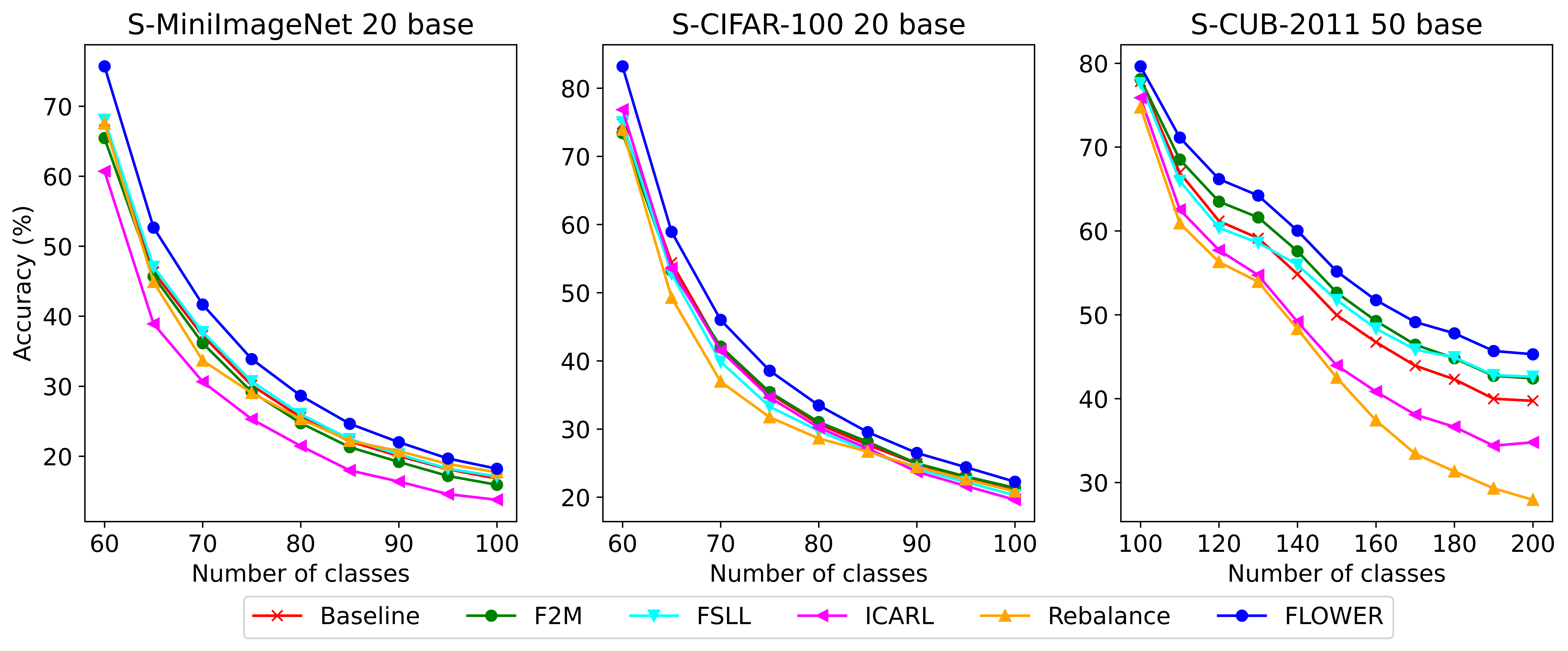}
 \caption{Visualization of Consolidated algorithms in MiniImagenet, CIFAR-100, and CUB-2011 dataset with 20, 20, and 50 base classes respectively.}
 \label{fig_small_base_visualization}
 \end{figure}

\begin{figure}[!ht]
 \centering
 \includegraphics[scale=0.28]{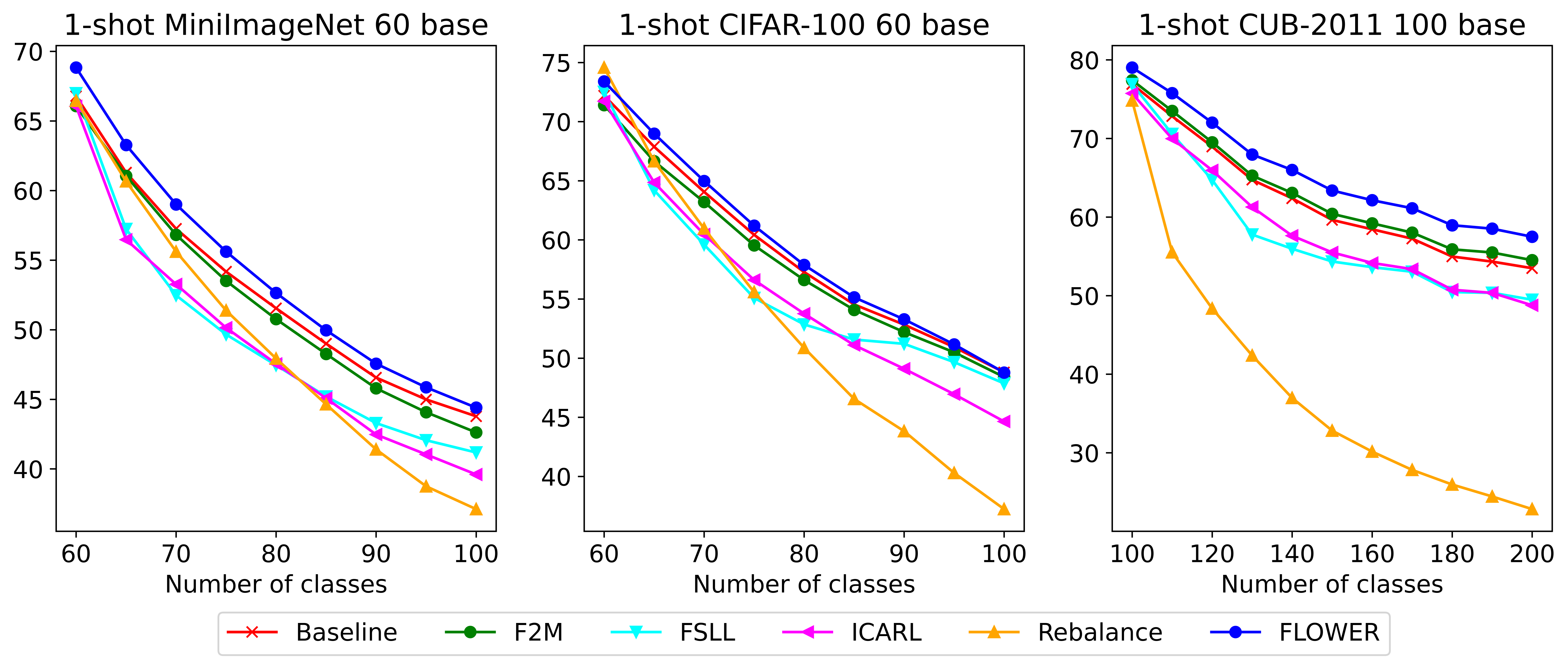}
 \caption{Visualization of Consolidated algorithms in MiniImagenet, CIFAR-100, and CUB-2011 dataset with 60, 60, and 100 base classes respectively with 1-shot setting.}
 \label{1shot_visualization}
 \end{figure}

Summarized numerical results in the 1-shot configuration are tabulated in Table \ref{tab:1shot}, while the detailed (per-session) results are tabulated in Table   \ref{tab:1shot_miniimagenet}, \ref{tab:1shot_cifar100}, and \ref{tab:1shot_cub}. FLOWER remains superior to other algorithms in the three benchmark problems under the 1-shot setting, i.e., each class of few-shot learning phases comprises 1 sample only. It produces better accuracy than baseline with visible gaps, i.e., $2\%$ for miniimagenet and CUB while its accuracy is on par to Rebalance in the CIFAR100. Note that the moderate base tasks are implemented here and Rebalance is poor on the CUB. Performance of consolidated algorithms for moderate base tasks under the 5-shot setting are pictorially illustrated in Fig. \ref{Fig:performance_plot} where ours consistently delivers the most improved trends. Visualizations for small base tasks under the 5-shot setting and that of the 1-shot setting are provided in Fig. \ref{fig_small_base_visualization} and Fig. \ref{1shot_visualization} where both exhibit the same and consistent trends .

\subsection{Ablation Study}
Ablation study is executed to study the contribution of each learning module of FLOWER to the overall learning performance where numerical results under the moderate base task of the miniimagenet problem are tabulated in Table \ref{tab:ablation}. In a nutshell, each learning module of FLOWER contributes positively where the absence of one component results in noticeable performance degradation. FLOWER is underpinned by the ball generator for feature space data augmentation to alleviate the issue of data scarcity where its absence significantly brings down FLOWER's accuracy by $13.70\%$. The absence of flat learning and wide learning incurs performance losses of $0.36\%$ and $0.5\%$ respectively.    

\subsection{Sensitivity Analysis}
Sensitivity analyses are performed to study the effect of the flat-wide region bound $b$ and the number of trials $M$. Numerical results across different bounds $b$ are reported in Table \ref{tab:bound value}. It is perceived that a too small bound leads to performance deterioration because of non-flat-wide regions, i.e., the confirmation of the presence of flat-wide local minima. On the other hand, a too large bound causes inaccurate approximations of flat-wide regions, thus ending up with the CF problem. $b=0.01$ is applied in our experiments. 

Numerical results across different random times $M$ are presented in Table \ref{tab:random_times}. $M=1$ significantly compromises FLOWER's performances since no approximation of flat-wide regions is carried out in this case. The increases of the number of trials $M>1$ result in trivial differences in performances. However, high $M$ slows down execution times since FLOWER has to undergo many trials in the base learning tasks to find wide-flat regions. In this paper, we set the number of trials $M\in[2,4]$ for all experiments. 

\section{Conclusion}
This paper presents FLOWER as a solution of few-shot continual learning problems. Our in-depth numerical study confirms the advantage of FLOWER where it outperforms prior arts with $1-4\%$ margin in the three benchmark problems for both average and per session accuracy. This fact is even more obvious with small base tasks where larger margins to its counterparts are observed than moderate base tasks whereas prior arts are compromised in such setting because they heavily focus on stability of base tasks. Ablation study and sensitivity analysis further demonstrates the advantage of each learning module and robustness of FLOWER. Current few-shot continual learning relies on a strong assumption where each task is drawn from the same domain. Our future study will be devoted to address cross-domain few-shot continual learning problems.

\bibliographystyle{IEEEtran}
\bibliography{ref-bibliography}

\begin{thebibliography}{10}
\providecommand{\url}[1]{#1}
\csname url@samestyle\endcsname
\providecommand{\newblock}{\relax}
\providecommand{\bibinfo}[2]{#2}
\providecommand{\BIBentrySTDinterwordspacing}{\spaceskip=0pt\relax}
\providecommand{\BIBentryALTinterwordstretchfactor}{4}
\providecommand{\BIBentryALTinterwordspacing}{\spaceskip=\fontdimen2\font plus
\BIBentryALTinterwordstretchfactor\fontdimen3\font minus
  \fontdimen4\font\relax}
\providecommand{\BIBforeignlanguage}[2]{{%
\expandafter\ifx\csname l@#1\endcsname\relax
\typeout{** WARNING: IEEEtran.bst: No hyphenation pattern has been}%
\typeout{** loaded for the language `#1'. Using the pattern for}%
\typeout{** the default language instead.}%
\else
\language=\csname l@#1\endcsname
\fi
#2}}
\providecommand{\BIBdecl}{\relax}
\BIBdecl

\bibitem{Parisi2019ContinualLL}
G.~I. Parisi, R.~Kemker, J.~L. Part, C.~Kanan, and S.~Wermter, ``Continual
  lifelong learning with neural networks: A review,'' \emph{Neural networks :
  the official journal of the International Neural Network Society}, vol. 113,
  pp. 54--71, 2019.

\bibitem{Ven2019ThreeSF}
G.~M. van~de Ven and A.~Tolias, ``Three scenarios for continual learning,''
  \emph{ArXiv}, vol. abs/1904.07734, 2019.

\bibitem{Finn2017ModelAgnosticMF}
C.~Finn, P.~Abbeel, and S.~Levine, ``Model-agnostic meta-learning for fast
  adaptation of deep networks,'' \emph{ArXiv}, vol. abs/1703.03400, 2017.

\bibitem{Tao2020FewShotCL}
X.~Tao, X.~Hong, X.~Chang, S.~Dong, X.~Wei, and Y.~Gong, ``Few-shot
  class-incremental learning,'' \emph{2020 IEEE/CVF Conference on Computer
  Vision and Pattern Recognition (CVPR)}, pp. 12\,180--12\,189, 2020.

\bibitem{Mazumder2021FewShotLL}
P.~Mazumder, P.~Singh, and P.~Rai, ``Few-shot lifelong learning,'' in
  \emph{AAAI}, 2021.

\bibitem{Chen2021IncrementalFL}
K.~Chen and C.-G. Lee, ``Incremental few-shot learning via vector quantization
  in deep embedded space,'' in \emph{ICLR}, 2021.

\bibitem{Zhang2021FewShotIL}
C.~Zhang, N.~Song, G.~Lin, Y.~Zheng, P.~Pan, and Y.~Xu, ``Few-shot incremental
  learning with continually evolved classifiers,'' \emph{2021 IEEE/CVF
  Conference on Computer Vision and Pattern Recognition (CVPR)}, pp.
  12\,450--12\,459, 2021.

\bibitem{Shi2021OvercomingCF}
G.~Shi, J.~Chen, W.~Zhang, L.-M. Zhan, and X.-M. Wu, ``Overcoming catastrophic
  forgetting in incremental few-shot learning by finding flat minima,'' in
  \emph{NeurIPS}, 2021.

\bibitem{Snell2017PrototypicalNF}
J.~Snell, K.~Swersky, and R.~S. Zemel, ``Prototypical networks for few-shot
  learning,'' \emph{ArXiv}, vol. abs/1703.05175, 2017.

\bibitem{Cha2021CPRCR}
S.~Cha, H.~Hsu, F.~du~Pin~Calmon, and T.~Moon, ``Cpr: Classifier-projection
  regularization for continual learning,'' \emph{ArXiv}, vol. abs/2006.07326,
  2021.

\bibitem{Sun2021MEDAMW}
P.~Sun, Y.~Ouyang, W.~Zhang, and X.~Dai, ``Meda: Meta-learning with data
  augmentation for few-shot text classification,'' in \emph{IJCAI}, 2021.

\bibitem{kirkpatrick2016overcoming}
\BIBentryALTinterwordspacing
J.~Kirkpatrick, R.~Pascanu, N.~Rabinowitz, J.~Veness, G.~Desjardins, A.~A.
  Rusu, K.~Milan, J.~Quan, T.~Ramalho, A.~Grabska-Barwinska, D.~Hassabis,
  C.~Clopath, D.~Kumaran, and R.~Hadsell, ``Overcoming catastrophic forgetting
  in neural networks,'' 2016, cite arxiv:1612.00796. [Online]. Available:
  \url{http://arxiv.org/abs/1612.00796}
\BIBentrySTDinterwordspacing

\bibitem{Zenke2017ContinualLT}
F.~Zenke, B.~Poole, and S.~Ganguli, ``Continual learning through synaptic
  intelligence,'' \emph{Proceedings of machine learning research}, vol.~70, pp.
  3987--3995, 2017.

\bibitem{Aljundi2018MemoryAS}
R.~Aljundi, F.~Babiloni, M.~Elhoseiny, M.~Rohrbach, and T.~Tuytelaars, ``Memory
  aware synapses: Learning what (not) to forget,'' in \emph{ECCV}, 2018.

\bibitem{Paik2020OvercomingCF}
I.~Paik, S.~Oh, T.~Kwak, and I.~Kim, ``Overcoming catastrophic forgetting by
  neuron-level plasticity control,'' \emph{ArXiv}, vol. abs/1907.13322, 2020.

\bibitem{Mao2021ContinualLV}
F.~Mao, W.~Weng, M.~Pratama, and E.~Yapp, ``Continual learning via inter-task
  synaptic mapping,'' \emph{ArXiv}, vol. abs/2106.13954, 2021.

\bibitem{Li2018LearningWF}
Z.~Li and D.~Hoiem, ``Learning without forgetting,'' \emph{IEEE Transactions on
  Pattern Analysis and Machine Intelligence}, vol.~40, pp. 2935--2947, 2018.

\bibitem{Rusu2016ProgressiveNN}
A.~A. Rusu, N.~C. Rabinowitz, G.~Desjardins, H.~Soyer, J.~Kirkpatrick,
  K.~Kavukcuoglu, R.~Pascanu, and R.~Hadsell, ``Progressive neural networks,''
  \emph{ArXiv}, vol. abs/1606.04671, 2016.

\bibitem{Yoon2018LifelongLW}
J.~Yoon, E.~Yang, J.~Lee, and S.~J. Hwang, ``Lifelong learning with dynamically
  expandable networks,'' \emph{ArXiv}, vol. abs/1708.01547, 2018.

\bibitem{Zoph2017NeuralAS}
B.~Zoph and Q.~V. Le, ``Neural architecture search with reinforcement
  learning,'' \emph{ArXiv}, vol. abs/1611.01578, 2017.

\bibitem{Li2019LearnTG}
X.~lai Li, Y.~Zhou, T.~Wu, R.~Socher, and C.~Xiong, ``Learn to grow: A
  continual structure learning framework for overcoming catastrophic
  forgetting,'' in \emph{ICML}, 2019.

\bibitem{Xu2021AdaptivePC}
J.~Xu, J.~Ma, X.~Gao, and Z.~Zhu, ``Adaptive progressive continual learning.''
  \emph{IEEE transactions on pattern analysis and machine intelligence},
  vol.~PP, 2021.

\bibitem{Ashfahani2022UnsupervisedCL}
A.~Ashfahani and M.~Pratama, ``Unsupervised continual learning in streaming
  environments,'' \emph{IEEE transactions on neural networks and learning
  systems}, vol.~PP, 2022.

\bibitem{Pratama2021UnsupervisedCL}
M.~Pratama, A.~Ashfahani, and E.~Lughofer, ``Unsupervised continual learning
  via self-adaptive deep clustering approach,'' \emph{ArXiv}, vol.
  abs/2106.14563, 2021.

\bibitem{Rakaraddi2022ReinforcedCL}
A.~Rakaraddi, S.-K. Lam, M.~Pratama, and M.~V. de~Carvalho, ``Reinforced
  continual learning for graphs,'' \emph{Proceedings of the 31st ACM
  International Conference on Information \& Knowledge Management}, 2022.

\bibitem{Rebuffi2017iCaRLIC}
S.-A. Rebuffi, A.~Kolesnikov, G.~Sperl, and C.~H. Lampert, ``icarl: Incremental
  classifier and representation learning,'' \emph{2017 IEEE Conference on
  Computer Vision and Pattern Recognition (CVPR)}, pp. 5533--5542, 2017.

\bibitem{LopezPaz2017GradientEM}
D.~Lopez-Paz and M.~Ranzato, ``Gradient episodic memory for continual
  learning,'' in \emph{NIPS}, 2017.

\bibitem{Chaudhry2019EfficientLL}
A.~Chaudhry, M.~Ranzato, M.~Rohrbach, and M.~Elhoseiny, ``Efficient lifelong
  learning with a-gem,'' \emph{ArXiv}, vol. abs/1812.00420, 2019.

\bibitem{Chaudhry2021UsingHT}
A.~Chaudhry, A.~Gordo, P.~Dokania, P.~H.~S. Torr, and D.~Lopez-Paz, ``Using
  hindsight to anchor past knowledge in continual learning,'' in \emph{AAAI},
  2021.

\bibitem{Buzzega2020DarkEF}
P.~Buzzega, M.~Boschini, A.~Porrello, D.~Abati, and S.~Calderara, ``Dark
  experience for general continual learning: a strong, simple baseline,''
  \emph{ArXiv}, vol. abs/2004.07211, 2020.

\bibitem{Dam2022ScalableAO}
T.~Dam, M.~Pratama, M.~M. Ferdaus, S.~G. Anavatti, and H.~Abbas, ``Scalable
  adversarial online continual learning,'' \emph{ArXiv}, vol. abs/2209.01558,
  2022.

\bibitem{VinciusdeCarvalho2022ClassIncrementalLV}
M.~V. de~Carvalho, M.~Pratama, J.~Zhang, and Y.~San, ``Class-incremental
  learning via knowledge amalgamation,'' \emph{ArXiv}, vol. abs/2209.02112,
  2022.

\bibitem{Hou2019LearningAU}
S.~Hou, X.~Pan, C.~C. Loy, Z.~Wang, and D.~Lin, ``Learning a unified classifier
  incrementally via rebalancing,'' \emph{2019 IEEE/CVF Conference on Computer
  Vision and Pattern Recognition (CVPR)}, pp. 831--839, 2019.

\bibitem{Kang2020DecouplingRA}
B.~Kang, S.~Xie, M.~Rohrbach, Z.~Yan, A.~Gordo, J.~Feng, and Y.~Kalantidis,
  ``Decoupling representation and classifier for long-tailed recognition,''
  \emph{ArXiv}, vol. abs/1910.09217, 2020.

\bibitem{Castro2018EndtoEndIL}
F.~M. Castro, M.~J. Mar{\'i}n-Jim{\'e}nez, N.~G. Mata, C.~Schmid, and
  A.~Karteek, ``End-to-end incremental learning,'' \emph{ArXiv}, vol.
  abs/1807.09536, 2018.

\end{thebibliography}

\end{document}